\documentclass{article}

\usepackage{PRIMEarxiv}

\usepackage[utf8]{inputenc} % allow utf-8 input
\usepackage[T1]{fontenc}    % use 8-bit T1 fonts
\usepackage{hyperref}       % hyperlinks
\usepackage{url}            % simple URL typesetting
\usepackage{booktabs}       % professional-quality tables
\usepackage{amsfonts}       % blackboard math symbols
\usepackage{nicefrac}       % compact symbols for 1/2, etc.
\usepackage{microtype}      % microtypography
\usepackage{lipsum}
\usepackage{fancyhdr}       % header
\usepackage{graphicx}       % graphics
\graphicspath{{media/}}     % organize your images and other figures under media/ folder

\usepackage{cite}
\usepackage{amsmath,amssymb,amsfonts}
\usepackage{algorithmic}
\usepackage{graphicx}
\usepackage{pbox}
\usepackage{textcomp}
\usepackage{xcolor}
\usepackage{soul}
\usepackage{url}
\usepackage{hyperref}
\usepackage[usestackEOL]{stackengine}
\strutlongstacks{T}
\usepackage{float}

\title{ \vspace{1.0cm} Violence Detection in Videos using Deep Recurrent and Convolutional Neural Networks\\
\thanks{
Thanks to the Natural Sciences and Engineering Research Council of Canada (NSERC), funding reference number RGPIN-2018-06233. } }

\author{
  Abdarahmane Traor\'e and Moulay A. Akhloufi, \textit{Senior Member IEEE}\\
  \textit{Perception, Robotics, and Intelligent Machines Research Group (PRIME)} \\
  \textit{Department of Computer Science, Universit\'e de Moncton}\ \\
  Moncton, NB, Canada\\
  \textit{{eat4651, moulay.akhloufi\}@umoncton.ca}} \\
}

\begin{document}
\maketitle

\begin{abstract}
Violence and abnormal behavior detection research have known an increase of interest in recent years, due mainly to a rise in crimes in large cities worldwide.
In this work, we propose a deep learning architecture for violence detection which combines both recurrent neural networks (RNNs) and 2-dimensional convolutional neural networks (2D CNN). In addition to video frames, we use optical flow computed using the captured sequences. 
CNN extracts spatial characteristics  in each frame, while RNN extracts temporal characteristics.
The use of optical flow allows to encode the movements in the scenes. 
The proposed approaches reach the same level as the state-of-the-art techniques and sometime surpass them.  It was validated on 3 databases achieving good results.
\end{abstract}

% keywords can be removed

\section{Introduction}
With a growing population and expanding cities, we are facing an unprecedented rise of criminality \cite{mt_increasing_2016}. 
Monitoring systems where a human watches multiple screens to detect violence, theft, or other abnormal behaviors are becoming obsolete. 
It is hard for persons to focus for a long time to detect violence when they must monitor large crowds.
The developed computer vision methods are less effective because of the large volume of data that must be processed. Indeed characteristics resulting from those methods are extracted manually, processed and then classified by an algorithm. This extraction takes time and becomes almost impossible to perform with a large dataset. With the progress in artificial intelligence, several methods have been developed to detect violence. In fact it is possible to train convolution neural networks (CNN) to extract spatial and temporal features from a video to classify its content. 
In this work, we introduce, end-to-end deep learning methods using RGB frames and an optical flow with a CNN-LSTM network to detect violent scenes in videos. The architectures presented reach the same level as modern techniques and sometime surpass them. Our approaches have been tested on 3 public databases to validate their performance.

\section{Related works}

Many methods of violence detection have been proposed in recent years. We can classify these techniques into two categories, classical machine learning and deep learning. \newline

\subsection{Machine learning approach}

Machine learning methods are based on algorithms such as  k-nearest neighbors (KNNs) \cite{altman_introduction_1992}, support vector machine (SVMs) \cite{noauthor_ieee_nodate}, or random forest. In addition to the popular descriptors like MoSIFT \cite{xu_violent_2014} or VIF \cite{hassner_violent_2012} to extract features, various extraction methods have been developed to detect violence.
For the detection of violence, some approaches use motion blobs. In \cite{gracia_fast_2015}, Gracia \textit{et al.} assume that violence scenes have a position and a shape. The method consists of computing the difference between two consecutive frames then binarize the result to obtain the number of motion blobs. The largest motion blobs are marked on a fight sequence and a no fight sequence. Only k motion blobs are selected. The k blobs are categorized by parameters such as centroid, area, or perimeter between blobs. The selected k blobs are then classified using SVMs, KNNs, and random forests. Motions blobs outperform methods like MoSIFT, SIFT, VIF, and LMP in term of accuracy on popular datasets.
In \cite{senst_crowd_2017}, Senst \textit{et al.} proposed a specialized method based on the Lagrangian theory to automatically detect violent scenes. A new spatio-temporal model based on the Lagrangian direction fields was used to capture the features. This new model exploits motion background compensation, appearances, and long-term motion information. The experiment was conducted on three databases, Hockey dataset, movies dataset, and violent crowd dataset. The new Lagrangian model trained with an SVM performs better than methods such as ViF and HOG + BoW.
In \cite{gao_violence_2016}, Gao \textit{et al.} proposed Oriented Violent Flows (OViF), a statistical motion orientations that takes full advantage
of the motion magnitude change information. AdaBoost is used to choose the features that are then trained by an SVM. Tests have been performed on Hockey dataset, and Violent Flow, and the results are superior to those of baselines such as LTP and ViF.
In \cite{xia_real_2018}, Xia \textit{et al.} present a method that uses a bi-channels CNN and an SVM. First, bi-channels CNN is used to extract two types of features. The first one represents the features of appearance, and the second one represents the difference between adjacent frames. The two features are then classified using SVM. The methods were tested on two databases, Hockey, and Violent crowd datasets. The results are superior to methods like HOG, HOF, MoSIFT, SIFT.

\subsection{Deep learning approach}

Methods of violence scenes detection with Deep Learning are generally based on Deep Convolutional Neural Network (DCNN).\newline
One approach to detect violence scenes is to use 3D CNN. These algorithms are computationally expensive but are accurate \cite{ding_violence_2014}, \cite{li_end_end_2018}, \cite{song_novel_2019}, \cite{Ullah2019}. For example,  Song \textit{et al.} \cite{song_novel_2019} propose the use of 3D CNN to extract both spatial and temporal features. The CNN is based on C3D introduced by Tran \textit{et al.} \cite{tran_learning_2015}. The CNN consists of eight 3D convolutional layers and five 3D pooling layers. They have also introduced a new frame sampling method based on the gray centroid to reduce frame redundancy caused by uniform sampling. This 3D CNN method has been validated on violence flow, Hockey dataset and movies dataset. In \cite{Ullah2019}, Ullah \textit{et al.} propose a three-stage deep learning approach for violence detection. First, a detection algorithm is used to track people in surveillance images and its output is feed to a 3D CNN where spatio-temporal features are extracted. The results of this 3D CNN are then sent to softmax classifier for violence detection. The 3D CNN is based on C3D \cite{tran_learning_2015} and is composed of eight convolutional and four pooling layers. It has been tested on Violent Flow, Hockey dataset and movies dataset. It ranks above the classical methods but is not the best of the Deep Learning methods.\newline
 There are also other algorithms that combine two types of neural networks (NN). For example, a 2D time distributed CNN in order to extract the spatio-temporal characteristics and an RNN to refine these characteristics. This end-to-end approach is computationally efficient and gives interesting results, as mentioned in \cite{abdali_robust_2019}. As features extractor Simonyan et al. use VGG19 \cite{simonyan_very_2015} pre-trained on ImageNet. The extracted features are fed to an LSTM whose output will be passed to a time distributed fully connected layer in order to detect violence. The approach was tested on Violent Flow, hochey dataset and movies dataset with results close to the state-of-the-art. Most algorithms which are based on CNN and RNN  use feature extraction followed by a Long Short-Term Memory (LSTM) \cite{hochreiter_long_1997}. There are 3 types of features extraction techniques, first the standard method that feed a whole frame to a CNN such as \cite{nguyen_violent_2019}, second a more sophisticated approache such as \cite{ditsanthia_video_2018} that divide a video frame into patches to extract characteristics from each patches using a CNN. This method of patch division allows to avoid the loss of discriminatory elements caused by differences in scale and location of persons in the frames. Finally technique that use CNNs which can provide multiscale features. In \cite{ditsanthia_video_2018}, Ditsanthia \textit{et al.} use a method called multiscale convolutional feature extraction to manage videos with different scales, and feeds the LSTMs with these multiscale extracted characteristics.\newline
 Some methods use other types of data in addition to RGB frames. Most often, it is optical flow. The optical flow makes it possible to encode the movement. Adding this information to the RGB frames allows to get better performence. To combine RGB and optical flow two subnetworks are trained, one on RGB frames and the other on optical flow, their outputs are then combined for classification, this the approach is used in \cite{xu_violent_2018}.\newline
 Finally, we have the methods that use special LSTMs called ConvLSTMs. These are LSTMs whose matrix operations have been replaced by convolutions. ConvLSTM enable to capture the temporal and spatial characteristics \cite{macintyre_detecting_2019},\cite{sudhakaran_learning_2017}. In \cite{sudhakaran_learning_2017}  Sudhakaran \textit{et al.} use this special convolution to classify actions. Features are first extracted by a CNN, aggregated using a ConvLSTM and then classified.

\section{Proposed Method}
\label{sec:method}

\begin{figure}[ht]
  \centering
 \includegraphics[width=8.5cm]{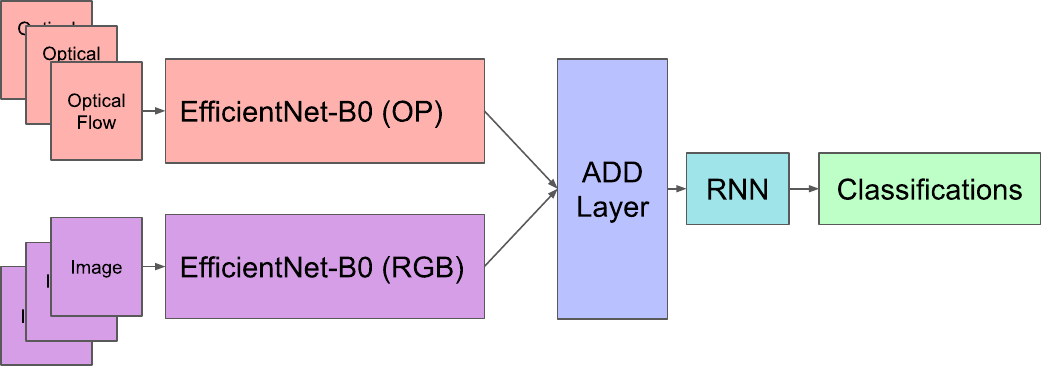}
\caption{Proposed architecture pipeline}
\label{figure:net}
\end{figure}

 We use 2D CNN, distributed in time to capture temporal and spatial characteristics. We combine this  2D CNN with a bi-directional
RNN (GRU or LSTM) to improve our detections. We also use optical flow to encode movement between frames for better performance. The network is composed of two identical specialized blocks (figure \ref{figure:net}), one for the RGB frames and the other for the optical flow. The characteristics from these two networks are then added together, refined by an RNN and then classified using a fully connected layer with sigmoid activation.

\subsection{Convolutional Neural Network}
\label{ssec:cnn}

\begin{figure}[ht]
\vspace{0.3cm}
  \centering
 \includegraphics[width=8.5cm]{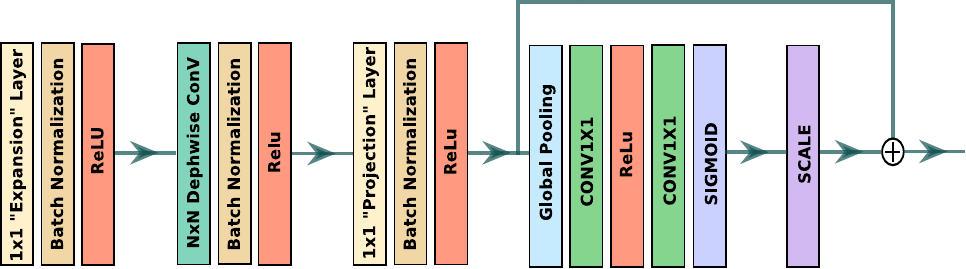}
\caption{MBCONV block of EfficientNet}
\label{figure:mbconv}
\end{figure}

\begin{figure}[ht]
  \centering
 \includegraphics[width=8.5cm]{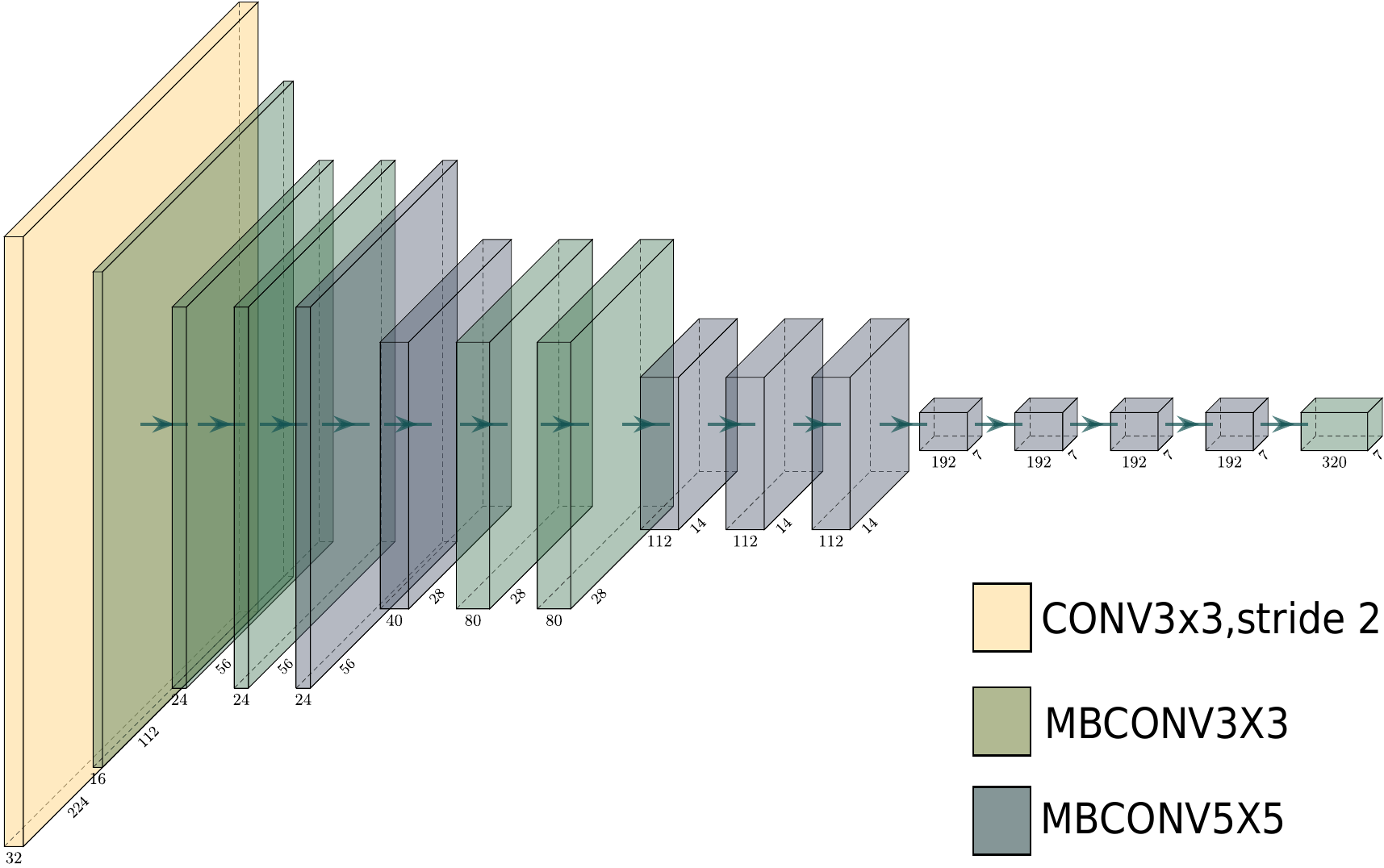}
\caption{EfficientNetB0 used to capture spatial features}
\label{figure:effb0}
\end{figure}

As a convolutional neural network, we have selected EfficientNet \cite{tan_efficientnet:_2019}, which is characterized by its compound scaling principle and its efficiency during inference. There are eight versions of EfficientNet (B0-B7),  the network consists of MBBLOCKS \cite{howard_mobilenets_2017} from MobileNets which are associated with a squeeze and excitation blocks \cite{hu_squeeze-and-excitation_2019}. Figure \ref{figure:mbconv} illustrate the MBCONV (MBBLOCK + squeeze and excitation block) used by EfficientNet. We used EfficientNet B0 (figure \ref{figure:effb0}) pre-trained on ImageNet \cite{imagenet_cvpr09}.

\subsection{Long Short-Term Memory}
\label{ssec:lstm}

The first RNN used to get a temporal representation is LSTM. This RNN has 3 gates: input gate, forget gate, and output gate. The input gate controls the amount of information that enters in the cell, the forget gate controls the flow of information that remains in the cell and the output gate controls the information that will be used to calculate the output activation of the LSTM unit. 
Equation \ref{eq:lstm} illustrates the LSTM used in this work \cite{gers_learning_2000}. $f_{t}$ represent the forget gate, $i_{t}$  the input gate, $o_{t}$ the output gate. $x_{t}$ is the input vector to the LSTM. $c_{t}$ and $\Tilde{c}_t$ are respectively the cell state vector and the cell input activation vector. $h_{t}$ is the hidden state vector, $W$ and $U$ are weights matrix that will be learned during the training and $b$ is the bias.$\sigma_{g}$ is a sigmoid activation function, and $\sigma_{h}$ is a hyperbolic tangent. Furthermore, we use our LSTM in a bidirectional position to consider not only the preceding sequences but also the following sequences, which helps to get better performance.

\begin{equation}
  \label{eq:lstm}
  \begin{gathered}
    f_{t} = \sigma_{g}({W_{f}x_{t} + U_{f}h_{t-1} + b_{f}}) \\
    i_{t} = \sigma_{g}({W_{i}x_{t} + U_{i}h_{t-1} + b_{i}}) \\
    o_{t} = \sigma_{g}({W_{o}x_{t} + U_{o}h_{t-1} + b_{o}}) \\
    \Tilde{c}_{t} = \sigma_{h}({W_{c}x_{t} + U_{c}h_{t-1} + b_{c}}) \\
    c_{t} = f_{t} \circ c_{t-1} +  i_{t} \circ \Tilde{c}_{t} \\
    h_{t} = o_{t} \circ \sigma_{h}(c_{t} )
  \end{gathered}
\end{equation}

\subsection{Gated Recurrent Unit}
\label{ssec:gru}

The GRU's option is made to prevent the LSTM's problem of gradient vanishing. For some tasks, GRU is more resistant to noise and outperforms the LSTM \cite{cho_learning_2014}. The GRUs are less computationally intensive because, unlike the LSTM which has three, they have two gates.

 Equation \ref{eq:gru} provides the GRU functions used in this work \cite{dey_gate-variants_2017}. $Z_{t}$, update gate determines what information to retain or drop, the $r_{t}$ reset gate determines how much past knowledge to forget, and $h_{t}$ is the output gate. W, U, and b are matrix and vector parameters, $\sigma_{g}$ is a sigmoid activation function, and $\sigma_{h}$ is a hyperbolic tangent. The input vector $x_{t}$. We use GRU in bidirectional mode to consider either following and preceding information.

\begin{equation}
  \label{eq:gru}
  \begin{gathered}
    Z_{t} = \sigma_{g}(W_{z}x_{t}+U_{z}h_{h-1}+b_{z}) \\
    r_{t} = \sigma_{g}(W_{r}x_{t}+U_{z}h_{h-1}+b_{z}) \\
    h_{t} = (1-z_{t}) \circ h_{t-1} + z_{t} \circ \sigma_{h}(W_{h}x_{t} +\\
    U_{h}(t_{t} \circ h_{t-1}) + b_{h}) 
  \end{gathered}
\end{equation}

\subsection{Optical Flow}
\label{ssec:flow}

\begin{figure}[ht]
    \centering
 \includegraphics[width=8.7cm]{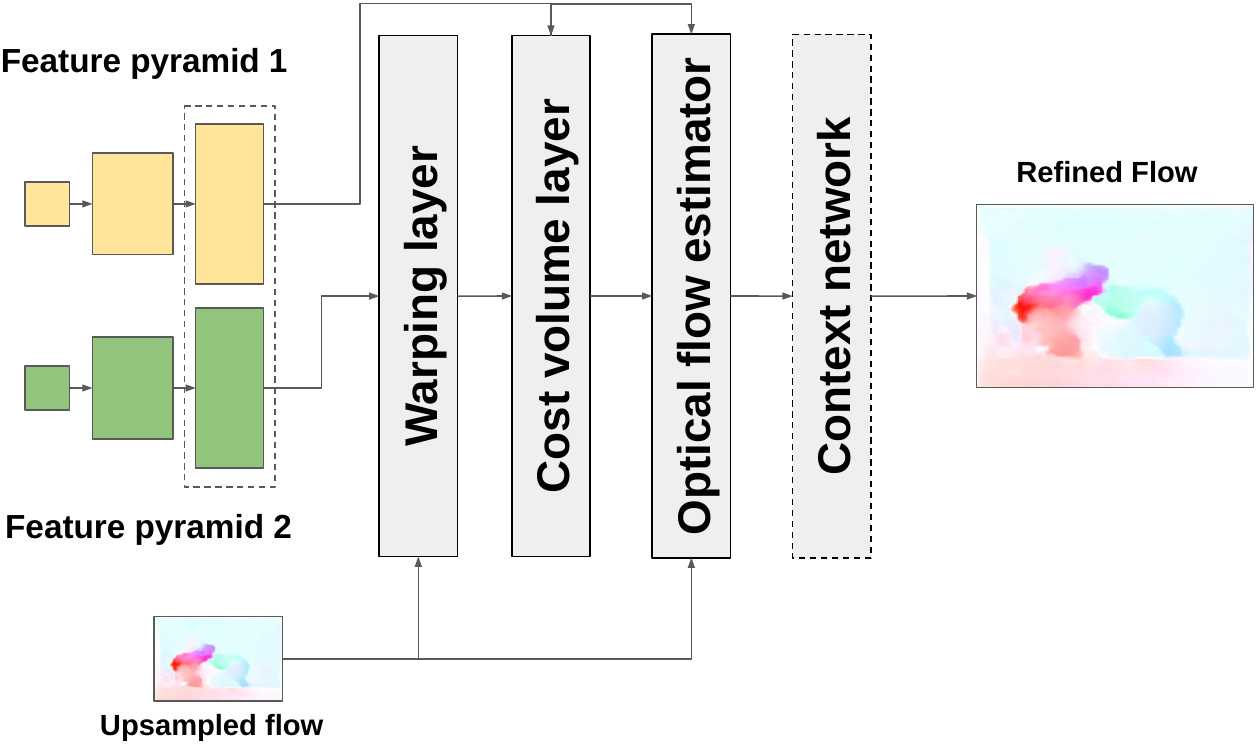}
 \caption{PWC-NET architecture}
    \label{fig:pwc}
\end{figure}

Optical flow is a method of perceiving movement in a sequence of images (videos) that can be calculated in different ways, in our case we opted for the use of PWC-Net \cite{sun_pwc-net_2018}. PWC-Net is an approach to obtain a smaller network than FLowNet2  \cite{ilg_flownet_2016} and also more efficient in term of accuracy by adding domain knowledge into the design of the network (see figure \ref{fig:pwc}). To compute the optical flow, PWC-Net first uses a learnable features pyramid to counteract the variations of shadows and brightness in the raw image. Then a warping operation is performed to capture large motions. After this warping operation the features are passed to a layer that compute the cost volume. Cost volume is a more discriminating representation of the optical flow than the image. The representation from the cost volume layer is then processed by CNN to estimate the flow. Since warping and cost volume have no learnable parameters, it reduces the size of the model. At the end of the pipeline a post-process of the context information is done by using a network to refine the optical flow. We used PWC-Net pre-trained on  MPI Sintel dataset \cite{Butler:ECCV:2012} to extract the optical flow for our datasets. 

\section{ Experiments and Results}
\label{sec:expares}

We used three datasets to test our network: Hockey dataset \cite{nievas2011violence}, Violent Flow dataset \cite{hassner_violent_2012}, and Real Life Crime situations dataset \cite{noauthor_real_nodate}. We used the accuracy metric to measure the performance of our network. 

\subsection{Datasets}
\label{ssec:datasets}

Datasets were randomly divided into two groups, training (80\%) and validation/testing (20\%). 

\subsubsection{Hockey Dataset} 
\label{sssec:Hockey}

\begin{figure}[ht]
\centering
\includegraphics[width=16cm,height=4cm]{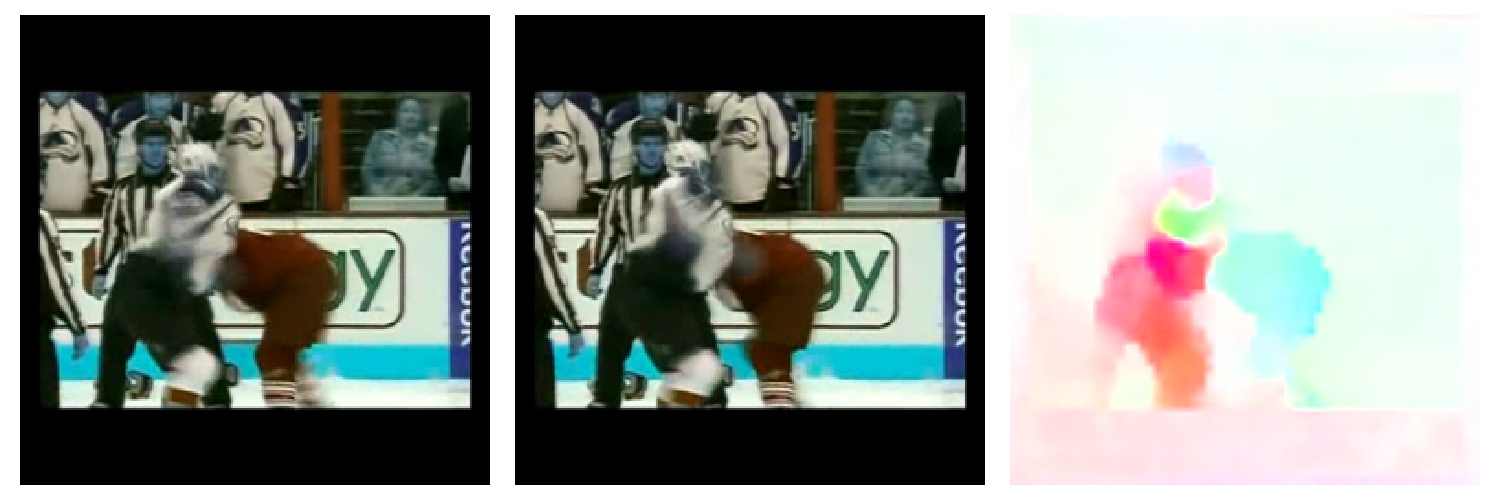}
\includegraphics[width=16cm,height=4cm]{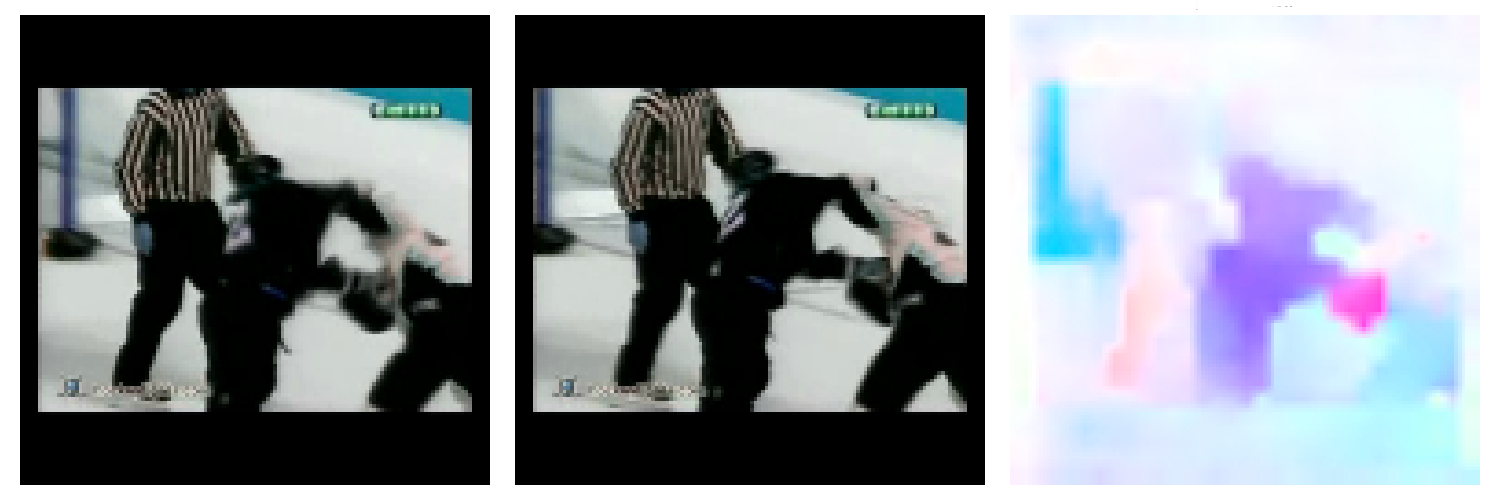}
\caption{Frames from Hockey Dataset, on the left we have the first frame, in the middle we have the second frame, on the right we have the optical flow compute using the first and the second frame.}
\label{fig:hockey}
\end{figure}

Hockey Fight Dataset \cite{nievas2011violence}  is a 2000 videos Dataset with 1000 fight and 1000 no fight videos of Hockey (figure \ref{fig:hockey}). Clips last approximately 2 seconds and consist of approximately 41 frames with a resolution of 360x288. The details of the sequences are quite similar. We have resized the clips frames to 128x128. 

\subsubsection{Violent Flow Dataset} 
\begin{figure}[ht]
\centering
\includegraphics[width=16cm,height=4cm]{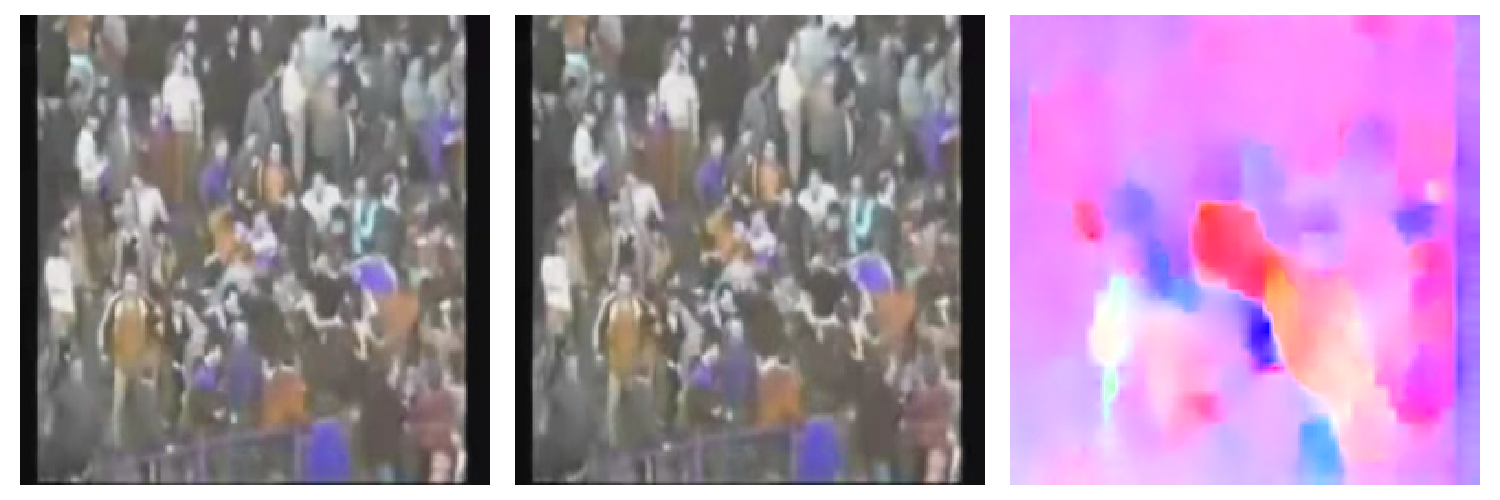}
\includegraphics[width=16cm,height=4cm]{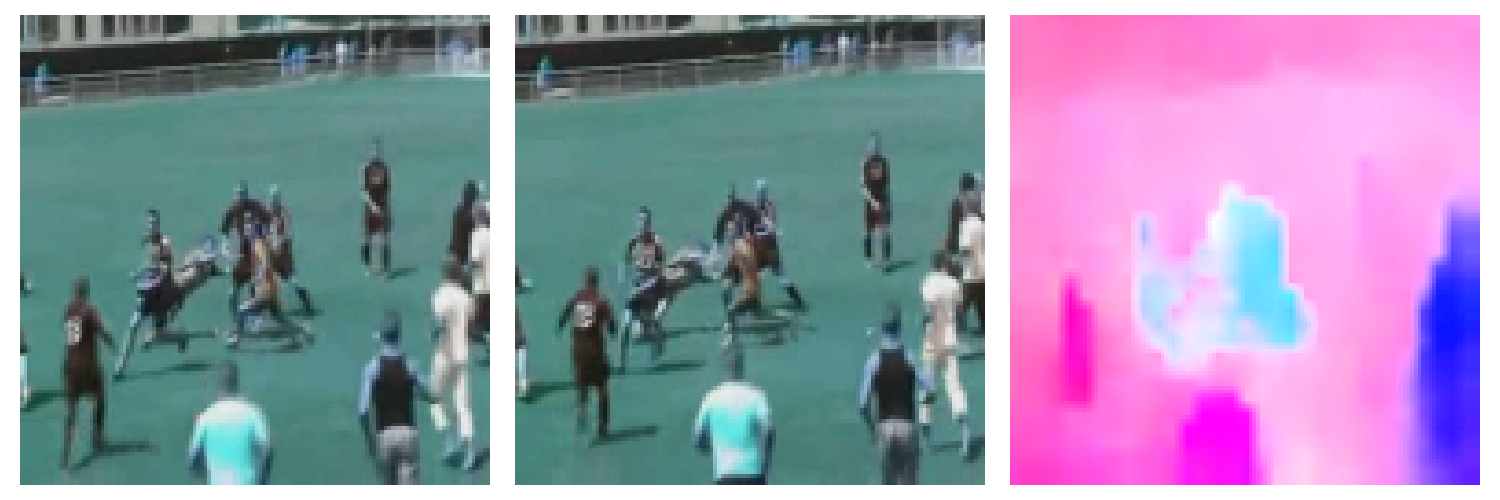}
\caption{Frames from ViolentFlow, on the left we have the first frame, in the middle we have the second frame, on the right we have the optical flow compute using the first and the second frame.}
\label{fig:flow}
\end{figure}

Violent Flow Dataset \cite{hassner_violent_2012}  is a dataset containing real-world video recordings of crowd violence (see figure \ref{fig:flow}). There are 246 videos in the dataset. The shortest clip length is 1.04 seconds, the longest clip is 6.52 seconds and the average length of the clip is 3.60 seconds. We have also resized the frames to 128x128. 

\subsubsection{Real Life Violence Situations Dataset} 
\label{sssec:rlviolence}

\begin{figure}[ht]
\vspace{2mm}
\centering
\includegraphics[width=16cm,height=4cm]{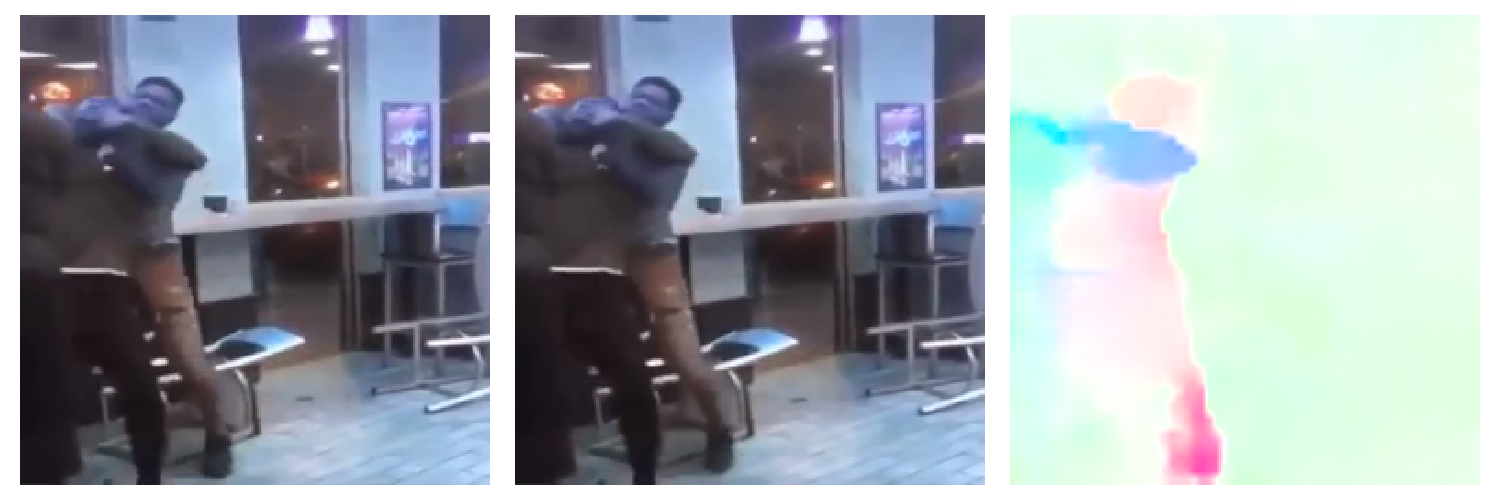}
\includegraphics[width=16cm,height=4cm]{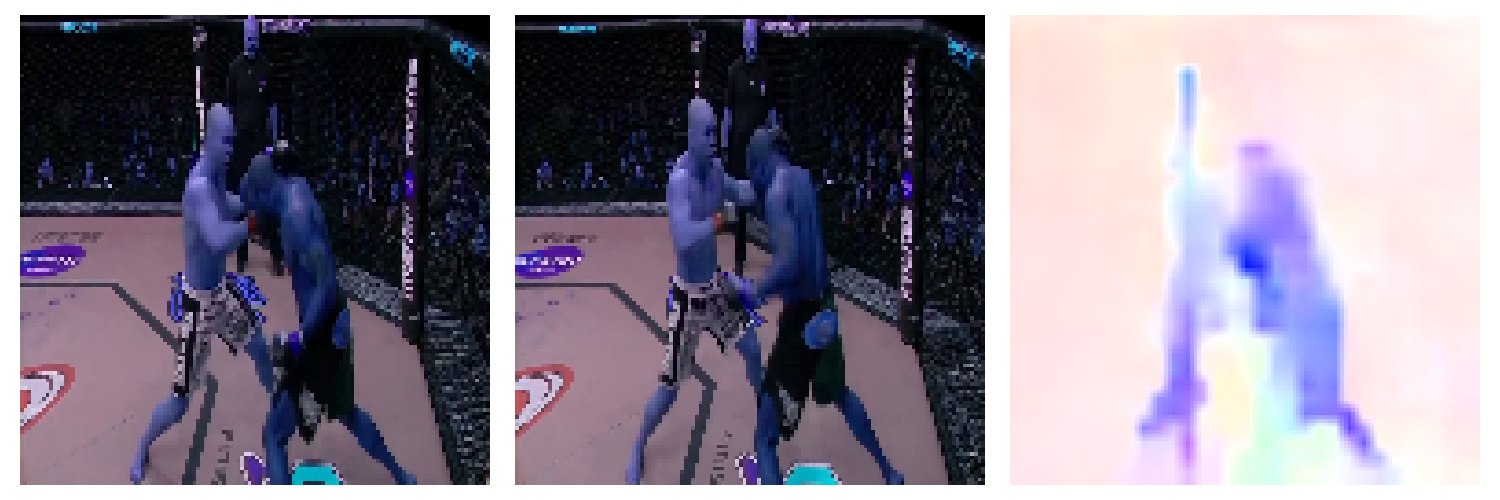}
\caption{Frames from Real Life Violence Situations Dataset, on the left we have the first frame, in the middle we have the second frame, on the right we have the optical flow compute using the first and the second frame.}
\label{fig:real}
\end{figure}

Real-Life Violence Situations Dataset \cite{noauthor_real_nodate} is a dataset of violence video clips from various situation of real life (see figure \ref{fig:real}). It contains 1000 fight and no fight sequence.
We resized the images to 128x128. 

\subsection{Parameters and sampling}
\label{ssec:settings}

We used Keras \cite{chollet2015keras} with the TensorFlow \cite{tensorflow2015-whitepaper} backend to set up our networks. 
Having different lengths of videos, we decided to choose a fixed number of 12 frames for each dataset. 
To choose these frames, we made a uniform sampling that allowed at the same time to avoid unnecessary computation of the network caused by redundant frames. For the computation of optical flow, we used  RGB frames that we have sampled and the frames that follow them. For example, for the computation of the optical flow for frame 6, we used frame 6 and frame 7. To go further in our experimentation, we have also made jumps of 2 and then 3 frames from the RGB frame, for example, for frame 6, we have also calculated the optical flow with frame 8 and frame 9.
Our EfficientNet-B0 has been pre-trained on ImageNet and then combined with RNNs, followed by 3 fully connected layers. We named this architecture ValdNet. There is three versions of ValdNet. ValdNet1,ValdNet2 and ValdNet3, the number in front of the name indicates the interval between frames when calculating the optical flow. Each version of ValdNet can have a GRU version and an LSTM version.

ValdNet was trained on all the datasets using rmsprop as  optimizer with a batch size of 4. The learning rate was set to 0.001. The networks were trained for 50 epochs.

\subsection{Results and discussions}
\label{ssec:result}

Tables \ref{table:hockey}, \ref{table:flow} and \ref{table:real}  present our results on the 3 datasets, Hockey, Violent Flow and Real Life Crime situations. The tables are composed of classic methods (Machine Learning) and Deep Learning methods. The classical methods have different inputs from frames like a bag of words or histograms bins.\newline
On Hockey dataset we obtained an accuracy of 99\% with ValdNet1 (GRU), results close to \cite{song_novel_2019} with 99.62\%. On Violent Flow, we obtained an accuracy of 93.53\% with ValdNet3 (LSTM), lower than our previous work with 95.00\% based on VGG+GRU  without optical flow. In the visualization of the optical flow for Violent Flow (figure \ref{fig:flow}) is not sharp enougth to help for the classification. On Real Life Crime situations dataset, we surpassed the result of our previous work with all versions of ValdNet. Our highest result is 96.74\% of accuracy, 6.24\% better than our previous experiment, and the other method in\cite{soliman_violence_2019}, which had 86.39\%.

\begin{table}[H]
\setlength{\tabcolsep}{0.1em}
\caption{Hockey Dataset violence detection, comparison with other methods (W represent words, O is optical flow, H represent histogram bins, F are frames, C is a CXC window, "?" means not specified and "-" means not used)}
\label{table:hockey}
\begin{center}
 \begin{tabular}{||l| c | c | c ||} 
 \hline
 Method & Inputs & Sampling & Accuracy (\%)  \\ [0.5ex] 
 \hline\hline
  AdaBoost + SVM \cite{gao_violence_2016} & 20 (H) & - & 87.50 \\
 \hline
 %HOF + BoW \cite{song_novel_2019} & 1000 (W) & - & 88.6\\ 
 %\hline
% HOG + BoW \cite{song_novel_2019} & 1000 (W) & - & 91.70\\
 %\hline
 %MoSIFT + BoW \cite{song_novel_2019}  & 1000 (W) & - & 90.90\\
 %\hline
%  MoWLD + BoW \cite{song_novel_2019}  & 900  (W) & - & 91.9\\
% \hline
 MoWLD + Sparse Coding \cite{song_novel_2019} & 1000 (W) & - & 93.70\\
 \hline
 \Longstack{MoSIFT + KDE + \\  Sparce Coding \cite{song_novel_2019}}  & 500 (W) & - & 94.30\\
 \hline
  LaSift \cite{senst_crowd_2017}   & 500 (W) & - & 94.42  \ \\
 \hline
 \Longstack{MoWLD + KDE + \\ Sparce Coding \cite{song_novel_2019}} & 500 (W) & - &   94.90\\
 \hline
 10.1109/TCSVT.2016.2589858 + KDE + SRC \cite{song_novel_2019}  & 1800 (W) & - & 96.80\\
 \hline
  3D-CNN \cite{song_novel_2019}  & 16 (F) & Uniform &    91.00\\
 \hline
Bi-channels CNN \cite{xia_real_2018}  & ? & Uniform &    95.90\\
\hline
 3D ConvNet \cite{Ullah2019} & 16 (F) & Intervall of 8 &  96.00\\
 \hline
  \underline{ValdNet2 (GRU)} & 12 (F+O) & Uniform &    96.00\\
 \hline
  FightNet \cite{fnet} & 25 (F) & Random &  97.00\\
 \hline
   VGG19+LSTM \cite{abdali_robust_2019} & 40 (F) & ? &  97.00\\
 \hline
 \underline{ValdNet3 (GRU)} & 12 (F+O) & Uniform &  97.00\\
 \hline
   ConvLSTM \cite{sudhakaran_learning_2017} & 20 (F) & Custom &  97.10\\
 \hline
   \underline{ValdNet1 (LSTM)} & 12 (F+O) &  Uniform &  98.00\\
 \hline
   VGG16 + GRU \cite{vgggru} & 10 (F) &  Uniform &  98.00\\
 \hline
  \underline{ValdNet2 (LSTM)} & 12 (F+O) &  Uniform &  98.00\\
 \hline
  \underline{ValdNet3 (LSTM)} & 12 (F+O) &  Uniform &  98.00\\
 \hline
  3D ConvNet \cite{song_novel_2019} & 16 (F) & Uniform & 98.96\\
 \hline
  \underline{\textbf{ValdNet1 (GRU)}} & 12 (F+O) &  Uniform &  \textbf{99.00}\\
  \hline
  \textbf{3D ConvNet}  \cite{song_novel_2019} & 32 (F) & Uniform & \textbf{99.62}\\
 \hline
\end{tabular}
\end{center}
\end{table}

\begin{table}[H]
\setlength{\tabcolsep}{0.1em}
\caption{Violent Flow Dataset violence detection, comparison with other methods (W, H, F, O, C, "?" and "-" are the same as in table \ref{table:hockey})}
\label{table:flow}
\begin{center}
 \begin{tabular}{||l | c | c | c||} 
 \hline
 Method & Inputs &  Sampling & Accuracy (\%)  \\ [0.5ex] 
 \hline\hline
% LTP \cite{song_novel_2019} & 20 (H)    & -  &71.53   \\ 
% \hline
% ViF \cite{song_novel_2019}  & 20 (H)  & -  & 81.30 \\
% \hline
 MoWLD + BoW \cite{song_novel_2019}  & 500 (W) & -  & 82.56 \\
 \hline
 RVD \cite{song_novel_2019}  & 4x4 (C) /5 frames & -  & 82.79 \\
 \hline
 MoWLD + Sparse Coding \cite{song_novel_2019}  & 500 (W) & -  & 86.39   \\
 \hline
 VGG19+LSTM \cite{abdali_robust_2019} & 40 (F) & ?  &  85.71 \\
 \hline
  AdaBoost + SVM \cite{gao_violence_2016} & 20 (H) & - & 88.00 \\ 
 \hline
 \Longstack{MMoSIFT + KDE + \\ Sparce Coding \cite{song_novel_2019}} & 500 (W) & -  & 89.05   \\
 \hline
 \Longstack{MMoWLD + KDE + \\ Sparce Coding \cite{song_novel_2019}}  & 500 (W) & -  &   89.78  \\
 \hline
\underline{ValdNet2 (GRU)} & 12 & Uniform  & 91.66 \\
\hline
\underline{ValdNet2 (LSTM)} & 12 & Uniform  & 91.66 \\
\hline
\underline{ValdNet1 (LSTM)} & 12 & Uniform  & 91.66 \\
\hline
 LaSift \cite{senst_crowd_2017}   & 500 (W) & - & 93.12   \\
 \hline
 MoWLD + KDE +SRC \cite{song_novel_2019} & 1800 (W) & -  &   93.19  \\
 \hline
Bi-channels CNN \cite{xia_real_2018}  & ? & Uniform &    93.25 \\
  \hline
3D ConvNet  \cite{song_novel_2019} & ? & Uniform  & 93.50\\
 \hline
\underline{ValdNet1 (GRU)} & 12 (F+O) & Uniform  & 93.75 \\
 \hline
\underline{ValdNet3 (GRU)} & 12 (F+O) & Uniform  & 93.75 \\
 \hline
 \underline{ValdNet3 (LSTM)} & 12 (F+O) & Uniform  & \textbf{93.75} \\
 \hline
ConvLSTM \cite{sudhakaran_learning_2017} & 20 (F) & ?  & 94.57 \\
\hline
VGG16 + GRU \cite{vgggru} & 10 (F) &  Uniform  & \textbf{95.50} \\
\hline
\end{tabular}
\end{center}
\end{table}

\begin{table}[ht]
\setlength{\tabcolsep}{0.1em}
\caption{Violent Flow Dataset violence detection, comparison with other methods (W, H, F, O, C, "?" and "-" are the same as in table \ref{table:hockey})}
\label{table:real}
\begin{center}
\begin{tabular}{||l | c | c | c||} 
\hline
Method & Inputs &  Sampling & Accuracy (\%)  \\ [0.5ex] 
\hline
VGG16 + LSTM \cite{soliman_violence_2019}  & ? (F) & -  & 86.39  \\
\hline
VGG16 + GRU \cite{vgggru} & 10 (F) &  Uniform  & 90.50 \\
\hline
\underline{ValdNet3 (LSTM)} & 12 (F + O) &  Uniform  & 93.75\\
\hline
\underline{ValdNet2 (LSTM)} & 12 (F + O) &  Uniform  & 93.99\\
\hline
\underline{ValdNet1 (GRU)}  & 12 (F + O)  &  Uniform  & 94.74\\
\hline
\underline{ValdNet3 (GRU)}  & 12 (F + O) &  Uniform  & 95.49\\
\hline
\underline{ValdNet1 (LSTM)} & 12 (F + O) &  Uniform  & 96.24\\
\hline
\underline{ValdNet2 (GRU)}  & 12 (F + O) &  Uniform  & \textbf{96.74}\\
\hline
\end{tabular}
\end{center}
\end{table}

\section{Conclusion and Future Works}

We used 2D  time distributed CNN  to capture spatio-temporal features, and we refined them using RNN. Also, we used two specialized sub-networks, one for RGB images and the other for optical flow, which outputs we have summed to encode the motion from the optical flow into our features. These combinations have allowed increasing our results in the proposed dataset.
We achieved 99\%, 93.75\%, and 96.74\% respectively on Hockey dataset, Violent Flow, and Real Life Crime Situations dataset. We are second on Hockey dataset behind \cite{song_novel_2019} by 0.62 difference. On ViolentFlow, the presence of several people seems to pose a problem to the optical flow. We know that there is movement but the low resolution of the optical flow does not allow to clearly determine what people are doing in the scene (figure \ref{fig:flow}). We could not achieve the result of our previous experiments without optical flow \cite{vgggru}. Real-Life Crime situations dataset is a fairly new database, so there is no other test for the moment except the baseline \cite{soliman_violence_2019}, which we outperformed by more than 10\%.
For our future work, we will introduce other datasets in order to benchmark all available databases and techniques. We will also benchmark the performance in terms of flops and speed of inference of all modern techniques available.

%Bibliography
\bibliographystyle{unsrt}  
\bibliography{references}

\begin{thebibliography}{10}

\bibitem{mt_increasing_2016}
Sheykhi MT.
\newblock Increasing {Crimes} vs. {Population} {Density} in {Megacities}.
\newblock {\em Sociology and Criminology-Open Access}, 4(1):1--2, 2016.

\bibitem{altman_introduction_1992}
N.~S. Altman.
\newblock An {Introduction} to {Kernel} and {Nearest}-{Neighbor}
  {Nonparametric} {Regression}.
\newblock {\em The American Statistician}, 46(3):175--185, August 1992.

\bibitem{noauthor_ieee_nodate}
Alexander J.~Smola Bernhard~Schölkopf.
\newblock {IEEE} {Xplore} {Book} {Abstract} - {Learning} with {Kernels}:
  {Support} {Vector} {Machines}, {Regularization}, {Optimization}, and
  {Beyond}, 2002.

\bibitem{xu_violent_2014}
Long Xu, Chen Gong, Jie Yang, Qiang Wu, and Lixiu Yao.
\newblock Violent video detection based on {MoSIFT} feature and sparse coding.
\newblock {\em 2014 IEEE International Conference on Acoustics, Speech and
  Signal Processing (ICASSP)}, pages 3538--3542, 2014.

\bibitem{hassner_violent_2012}
Tal Hassner, Yossi Itcher, and Orit Kliper-Gross.
\newblock Violent flows: {Real}-time detection of violent crowd behavior.
\newblock In {\em 2012 {IEEE} {Computer} {Society} {Conference} on {Computer}
  {Vision} and {Pattern} {Recognition} {Workshops}}, pages 1--6, June 2012.

\bibitem{gracia_fast_2015}
Ismael~Serrano Gracia, Oscar~Deniz Suarez, Gloria~Bueno Garcia, and Tae-Kyun
  Kim.
\newblock Fast {Fight} {Detection}.
\newblock {\em PLOS ONE}, 10(4):e0120448, April 2015.

\bibitem{senst_crowd_2017}
Tobias Senst, Volker Eiselein, Alexander Kuhn, and Thomas Sikora.
\newblock Crowd {Violence} {Detection} {Using} {Global} {Motion}-{Compensated}
  {Lagrangian} {Features} and {Scale}-{Sensitive} {Video}-{Level}
  {Representation}.
\newblock {\em IEEE Transactions on Information Forensics and Security},
  12(12):2945--2956, December 2017.

\bibitem{gao_violence_2016}
Yuan Gao, Hong Liu, Xiaohu Sun, Can Wang, and Yi~Liu.
\newblock Violence detection using {Oriented} {VIolent} {Flows}.
\newblock {\em Image and Vision Computing}, 48-49:37--41, April 2016.

\bibitem{xia_real_2018}
Qing Xia, Ping Zhang, JingJing Wang, Ming Tian, and Chun Fei.
\newblock Real {Time} {Violence} {Detection} {Based} on {Deep}
  {Spatio}-{Temporal} {Features}.
\newblock In Jie Zhou, Yunhong Wang, Zhenan Sun, Zhenhong Jia, Jianjiang Feng,
  Shiguang Shan, Kurban Ubul, and Zhenhua Guo, editors, {\em Biometric
  {Recognition}}, Lecture {Notes} in {Computer} {Science}, pages 157--165,
  Cham, 2018. Springer International Publishing.

\bibitem{ding_violence_2014}
Chunhui Ding, Shouke Fan, Ming Zhu, Weiguo Feng, and Baozhi Jia.
\newblock Violence {Detection} in {Video} by {Using} {3D} {Convolutional}
  {Neural} {Networks}.
\newblock In {\em {ISVC}}, 2014.

\bibitem{li_end_end_2018}
Chengyang Li, Liping Zhu, Dandan Zhu, Jiale Chen, Zhanghui Pan, Xue Li, and
  Bing Wang.
\newblock End-to-end {Multiplayer} {Violence} {Detection} {Based} on {Deep} 3d
  {CNN}.
\newblock In {\em Proceedings of the 2018 {VII} {International} {Conference} on
  {Network}, {Communication} and {Computing}}, {ICNCC} 2018, pages 227--230,
  New York, NY, USA, 2018. ACM.
\newblock event-place: Taipei City, Taiwan.

\bibitem{song_novel_2019}
Wei Song, Dongliang Zhang, Xiaobing Zhao, Jing Yu, Rui Zheng, and Antai Wang.
\newblock A {Novel} {Violent} {Video} {Detection} {Scheme} {Based} on
  {Modified} 3d {Convolutional} {Neural} {Networks}.
\newblock {\em IEEE Access}, 7:39172--39179, 2019.

\bibitem{Ullah2019}
Fath U.~Min Ullah, Amin Ullah, Khan Muhammad, Ijaz~Ul Haq, and Sung~Wook Baik.
\newblock Violence detection using spatiotemporal features with {3D}
  convolutional neural network.
\newblock {\em Sensors (Basel, Switzerland)}, 19(11):2472, May 2019.

\bibitem{tran_learning_2015}
Du~Tran, Lubomir Bourdev, Rob Fergus, Lorenzo Torresani, and Manohar Paluri.
\newblock Learning {Spatiotemporal} {Features} with {3D} {Convolutional}
  {Networks}.
\newblock {\em arXiv:1412.0767 [cs]}, October 2015.
\newblock arXiv: 1412.0767.

\bibitem{abdali_robust_2019}
Al-Maamoon~R. Abdali and Rana~F. Al-Tuma.
\newblock Robust {Real}-{Time} {Violence} {Detection} in {Video} {Using} {CNN}
  {And} {LSTM}.
\newblock In {\em 2019 2nd {Scientific} {Conference} of {Computer} {Sciences}
  ({SCCS})}, pages 104--108, March 2019.

\bibitem{simonyan_very_2015}
Karen Simonyan and Andrew Zisserman.
\newblock Very {Deep} {Convolutional} {Networks} for {Large}-{Scale} {Image}
  {Recognition}.
\newblock {\em arXiv:1409.1556 [cs]}, April 2015.
\newblock arXiv: 1409.1556.

\bibitem{hochreiter_long_1997}
Sepp Hochreiter and Jürgen Schmidhuber.
\newblock Long {Short}-{Term} {Memory}.
\newblock {\em Neural Computation}, 9(8):1735--1780, November 1997.

\bibitem{nguyen_violent_2019}
Shakil~Ahmed Sumon, MD.~Tanzil Shahria, MD.~Raihan Goni, Nazmul Hasan, A.~M.
  Almarufuzzaman, and Rashedur~M. Rahman.
\newblock Violent {Crowd} {Flow} {Detection} {Using} {Deep} {Learning}.
\newblock In Ngoc~Thanh Nguyen, Ford~Lumban Gaol, Tzung-Pei Hong, and Bogdan
  Trawiński, editors, {\em Intelligent {Information} and {Database}
  {Systems}}, volume 11431, pages 613--625. Springer International Publishing,
  Cham, 2019.

\bibitem{ditsanthia_video_2018}
Eknarin Ditsanthia, Luepol Pipanmaekaporn, and Suwatchai Kamonsantiroj.
\newblock Video {Representation} {Learning} for {CCTV}-{Based} {Violence}
  {Detection}.
\newblock In {\em 2018 3rd {Technology} {Innovation} {Management} and
  {Engineering} {Science} {International} {Conference} ({TIMES}-{iCON})}, pages
  1--5, December 2018.

\bibitem{xu_violent_2018}
Xingyu Xu, Xiaoyu Wu, Ge~Wang, and Huimin Wang.
\newblock Violent {Video} {Classification} {Based} on {Spatial}-{Temporal}
  {Cues} {Using} {Deep} {Learning}.
\newblock In {\em 2018 11th {International} {Symposium} on {Computational}
  {Intelligence} and {Design} ({ISCID})}, volume~01, pages 319--322, December
  2018.

\bibitem{macintyre_detecting_2019}
Giorgio Morales, Itamar Salazar-Reque, Joel Telles, and Daniel Díaz.
\newblock Detecting {Violent} {Robberies} in {CCTV} {Videos} {Using} {Deep}
  {Learning}.
\newblock In John MacIntyre, Ilias Maglogiannis, Lazaros Iliadis, and Elias
  Pimenidis, editors, {\em Artificial {Intelligence} {Applications} and
  {Innovations}}, volume 559, pages 282--291. Springer International
  Publishing, Cham, 2019.

\bibitem{sudhakaran_learning_2017}
S.~{Sudhakaran} and O.~{Lanz}.
\newblock Learning to detect violent videos using convolutional long short-term
  memory.
\newblock In {\em 2017 14th IEEE International Conference on Advanced Video and
  Signal Based Surveillance (AVSS)}, pages 1--6, Aug 2017.

\bibitem{tan_efficientnet:_2019}
Mingxing Tan and Quoc~V. Le.
\newblock {EfficientNet}: {Rethinking} {Model} {Scaling} for {Convolutional}
  {Neural} {Networks}.
\newblock {\em arXiv:1905.11946 [cs, stat]}, May 2019.
\newblock arXiv: 1905.11946.

\bibitem{howard_mobilenets_2017}
Andrew~G. Howard, Menglong Zhu, Bo~Chen, Dmitry Kalenichenko, Weijun Wang,
  Tobias Weyand, Marco Andreetto, and Hartwig Adam.
\newblock {MobileNets}: {Efficient} {Convolutional} {Neural} {Networks} for
  {Mobile} {Vision} {Applications}.
\newblock {\em arXiv:1704.04861 [cs]}, April 2017.
\newblock arXiv: 1704.04861.

\bibitem{hu_squeeze-and-excitation_2019}
Jie Hu, Li~Shen, Samuel Albanie, Gang Sun, and Enhua Wu.
\newblock Squeeze-and-{Excitation} {Networks}.
\newblock {\em arXiv:1709.01507 [cs]}, May 2019.
\newblock arXiv: 1709.01507.

\bibitem{imagenet_cvpr09}
J.~Deng, W.~Dong, R.~Socher, L.-J. Li, K.~Li, and L.~Fei-Fei.
\newblock {ImageNet: A Large-Scale Hierarchical Image Database}.
\newblock In {\em CVPR09}, 2009.

\bibitem{gers_learning_2000}
Felix~A. Gers, Jürgen Schmidhuber, and Fred Cummins.
\newblock Learning to {Forget}: {Continual} {Prediction} with {LSTM}.
\newblock {\em Neural Computation}, 12(10):2451--2471, October 2000.

\bibitem{cho_learning_2014}
Kyunghyun Cho, Bart van Merrienboer, Caglar Gulcehre, Dzmitry Bahdanau, Fethi
  Bougares, Holger Schwenk, and Yoshua Bengio.
\newblock Learning {Phrase} {Representations} using {RNN} {Encoder}-{Decoder}
  for {Statistical} {Machine} {Translation}.
\newblock {\em arXiv:1406.1078 [cs, stat]}, September 2014.
\newblock arXiv: 1406.1078.

\bibitem{dey_gate-variants_2017}
Rahul Dey and Fathi~M. Salem.
\newblock Gate-{Variants} of {Gated} {Recurrent} {Unit} ({GRU}) {Neural}
  {Networks}.
\newblock {\em arXiv:1701.05923 [cs, stat]}, January 2017.
\newblock arXiv: 1701.05923.

\bibitem{sun_pwc-net_2018}
Deqing Sun, Xiaodong Yang, Ming-Yu Liu, and Jan Kautz.
\newblock {PWC}-{Net}: {CNNs} for {Optical} {Flow} {Using} {Pyramid},
  {Warping}, and {Cost} {Volume}.
\newblock {\em arXiv:1709.02371 [cs]}, June 2018.
\newblock arXiv: 1709.02371.

\bibitem{ilg_flownet_2016}
Eddy Ilg, Nikolaus Mayer, Tonmoy Saikia, Margret Keuper, Alexey Dosovitskiy,
  and Thomas Brox.
\newblock {FlowNet} 2.0: {Evolution} of {Optical} {Flow} {Estimation} with
  {Deep} {Networks}.
\newblock {\em arXiv:1612.01925 [cs]}, December 2016.
\newblock arXiv: 1612.01925.

\bibitem{Butler:ECCV:2012}
D.~J. Butler, J.~Wulff, G.~B. Stanley, and M.~J. Black.
\newblock A naturalistic open source movie for optical flow evaluation.
\newblock In {A. Fitzgibbon et al. (Eds.)}, editor, {\em European Conf. on
  Computer Vision (ECCV)}, Part IV, LNCS 7577, pages 611--625. Springer-Verlag,
  October 2012.

\bibitem{nievas2011violence}
Enrique~Bermejo Nievas, Oscar~Deniz Suarez, Gloria~Bueno Garcia, and Rahul
  Sukthankar.
\newblock Hockey fight detection dataset.
\newblock In {\em Computer Analysis of Images and Patterns}, pages 332--339.
  Springer, 2011.

\bibitem{noauthor_real_nodate}
Elesawy Mohamed, Hussein Mohamad, and Mina Abd~El Massih.
\newblock Real {Life} {Violence} {Situations} {Dataset}.
\newblock
  https://kaggle.com/mohamedmustafa/real-life-violence-situations-dataset.

\bibitem{chollet2015keras}
Fran\c{c}ois Chollet et~al.
\newblock Keras.
\newblock \url{https://github.com/fchollet/keras}, 2015.

\bibitem{tensorflow2015-whitepaper}
Mart\'{\i}n Abadi, Ashish Agarwal, Paul Barham, Eugene Brevdo, Zhifeng Chen,
  Craig Citro, Greg~S. Corrado, Andy Davis, Jeffrey Dean, Matthieu Devin,
  Sanjay Ghemawat, Ian Goodfellow, Andrew Harp, Geoffrey Irving, Michael Isard,
  Yangqing Jia, Rafal Jozefowicz, Lukasz Kaiser, Manjunath Kudlur, Josh
  Levenberg, Dandelion Man\'{e}, Rajat Monga, Sherry Moore, Derek Murray, Chris
  Olah, Mike Schuster, Jonathon Shlens, Benoit Steiner, Ilya Sutskever, Kunal
  Talwar, Paul Tucker, Vincent Vanhoucke, Vijay Vasudevan, Fernanda Vi\'{e}gas,
  Oriol Vinyals, Pete Warden, Martin Wattenberg, Martin Wicke, Yuan Yu, and
  Xiaoqiang Zheng.
\newblock {TensorFlow}: Large-scale machine learning on heterogeneous systems,
  2015.
\newblock Software available from tensorflow.org.

\bibitem{soliman_violence_2019}
Mohamed~Mostafa Soliman, Mohamed~Hussein Kamal, Mina~Abd El-Massih~Nashed,
  Youssef~Mohamed Mostafa, Bassel~Safwat Chawky, and Dina Khattab.
\newblock Violence {Recognition} from {Videos} using {Deep} {Learning}
  {Techniques}.
\newblock In {\em 2019 {Ninth} {International} {Conference} on {Intelligent}
  {Computing} and {Information} {Systems} ({ICICIS})}, pages 80--85, December
  2019.

\bibitem{fnet}
Peipei Zhou, Qinghai Ding, Haibo Luo, and Xinglin Hou.
\newblock Violent interaction detection in video based on deep learning.
\newblock {\em Journal of Physics: Conference Series}, 844:012044, 06 2017.

\bibitem{vgggru}
Abdarahmane Traoré and Moulay~A. Akhloufi.
\newblock {2D} {Bidirectional} {Gated} {Recurrent} {Unit} {Convolutional}
  {Neural} {Networks} for {End}-to-{End} {Violence} {Detection} in {Videos}.
\newblock In Aurélio Campilho, Fakhri Karray, and Zhou Wang, editors, {\em
  Image {Analysis} and {Recognition}}, Lecture {Notes} in {Computer} {Science},
  pages 152--160, Cham, 2020. Springer International Publishing.

\end{thebibliography}

\end{document}